%% file: main.tex
\documentclass[10pt,twocolumn,letterpaper]{article}

\usepackage[pagenumbers]{iccv} 
\usepackage{multicol}
\usepackage{multirow}
\usepackage{graphicx}
\usepackage{overpic}
\usepackage{tikz}
\usepackage{nth}

\input{preamble}

\definecolor{iccvblue}{rgb}{0.21,0.49,0.74}
\usepackage[pagebackref,breaklinks,colorlinks,allcolors=iccvblue]{hyperref}

\def\paperID{3057} 
\def\confName{ICCV}
\def\confYear{2025}

\title{LIA-X: Interpretable Latent Portrait Animator}

\author{
Yaohui Wang\textsuperscript{\rm 1} \hskip 0.5em
Di Yang\textsuperscript{\rm 2} \hskip 0.5em
Xinyuan Chen\textsuperscript{\rm 1} \hskip 0.5em
François Brémond\textsuperscript{\rm 2} \hskip 0.5em
Yu Qiao\textsuperscript{\rm 1*} \hskip 0.5em
Antitza Dantcheva\textsuperscript{\rm 2}
 \\
\small \textsuperscript{1}Shanghai Artificial Intelligence Laboratory \hskip 1em 
\small \textsuperscript{2}Inria, Université Côte d'Azur \\
{\small \url{https://wyhsirius.github.io/LIA-X-project/}}
}

\begin{document}


\twocolumn[{%
\renewcommand\twocolumn[1][]{#1}%
\maketitle
\vspace{-3em}
\begin{center}
\centering
\captionsetup{type=figure}
\centering
\includegraphics[width=1.0\linewidth]{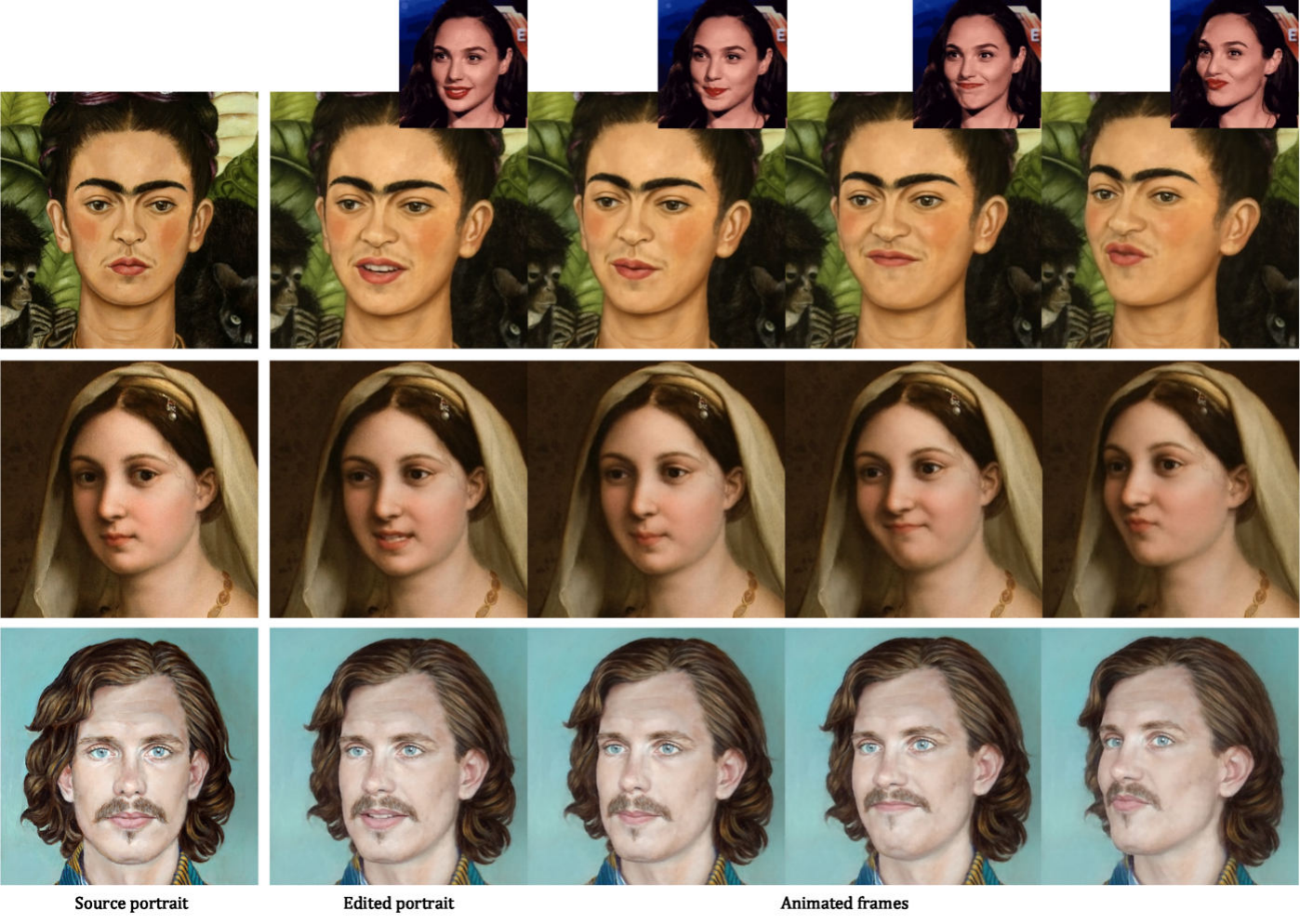}
\caption{\textbf{Portrait Animation.} We show results of three portraits animated by LIA-X using the video of Gal Gadot (top right). Given a source portrait ($1^{\text{st}}$ column), LIA-X allows for editing the source portrait based on the learned semantics to align it with the initial driving frame in terms of head pose and facial expression ($2^{\text{nd}}$ column). The animated sequences ($2^{\text{nd}}$-$5^{\text{th}}$ columns) are then obtained by applying motion transfer process on the edited portraits.}
\label{fig:teaser}
\end{center}%
%
}
]


\blfootnote{*Corresponding author.}

\input{sec/0_abstract}
\input{sec/1_intro}
\input{sec/2_related_work}
\input{sec/3_method}

\input{sec/4_experiments}
\input{sec/5_limitation}

\input{sec/6_conclusion}
{
    \small
    \bibliographystyle{ieeenat_fullname}
    \bibliography{main}
}

\end{document}

%% file: preamble.tex
\makeatletter
\newcommand\blfootnote[1]{%
  \begingroup
  \renewcommand\thefootnote{}%
  \def\@makefntext##1{\noindent##1}
  \footnote{#1}%
  \addtocounter{footnote}{-1}%
  \endgroup
}
\makeatother

%% file: sec/0_abstract.tex
\begin{abstract}
We introduce LIA-X, a novel interpretable portrait animator designed to transfer facial dynamics from a driving video to a source portrait with fine-grained control. LIA-X is an autoencoder that models motion transfer as a linear navigation of motion codes in latent space. Crucially, it incorporates a novel Sparse Motion Dictionary that enables the model to disentangle facial dynamics into interpretable factors. Deviating from previous 'warp-render' approaches, the interpretability of the Sparse Motion Dictionary allows LIA-X to support a highly controllable 'edit-warp-render' strategy, enabling precise manipulation of fine-grained facial semantics in the source portrait. This helps to narrow initial differences with the driving video in terms of pose and expression. Moreover, we demonstrate the scalability of LIA-X by successfully training a large-scale model with approximately 1 billion parameters on extensive datasets. Experimental results show that our proposed method outperforms previous approaches in both self-reenactment and cross-reenactment tasks across several benchmarks. Additionally, the interpretable and controllable nature of LIA-X supports practical applications such as fine-grained, user-guided image and video editing, as well as 3D-aware portrait video manipulation.

\end{abstract}


%% file: sec/1_intro.tex
\section{Introduction}
\label{sec:intro}
With the remarkable development of deep generative models~\cite{ddpm,ddim,goodfellow2014generative}, techniques for video generation have been largely advanced. Portrait animation, a domain-specific video generation task that aims to transfer facial dynamics from a driving video to a portrait image, has also received increasing attention due to its wide applications in entertainment, e-education, and digital human creation.

Towards accurately transferring facial dynamics, one straightforward strategy is to leverage pre-computed explicit representations such as facial landmarks~\cite{wang2019fewshotvid2vid, magicdance, ma2024followyouremoji}, 3DMM~\cite{liu2019liquid,Chen_2021_CVPR}, optical flows~\cite{li2018flow,ohnishi2018ftgan} and dense poses~\cite{xu2023omniavatar} as motion guidance. Such an approach has been widely used in both GAN-based and recent diffusion-based methods~\cite{magicdance, ma2024followyouremoji}. However, the generated quality of these methods relies heavily on the performance of the off-the-shelf feature extractors, which strongly restricts their usability in more challenging, real-world scenarios.

Self-supervised learning-based techniques have also been proposed to tackle this problem. Previous methods explored learning either explicit structures such as 2D/3D keypoints~\cite{fomm, siarohin2021motion, thinplate, facevid2vid, liveportrait} or implicit motion codes~\cite{lia-pami,wang2022latent} in an end-to-end manner to model facial dynamics. They typically follow a \textit{`warp-render'} strategy to conduct portrait animation based on the computed optical-flow fields. While these methods have achieved promising results in both motion transferring and identity preservation, their performance still drops significantly when large variations exist between the source and driving data in terms of head poses and facial expressions.

Towards addressing this issue, we introduce LIA-X, a novel framework that learns interpretable motion semantics to enable a highly controllable \textit{`edit-warp-render'} animation strategy. Deviating from the standard \textit{`warp-render'} pipeline, LIA-X allows users to first leverage the learned motion representations to align source portrait with the initial driving frame before applying the final motion transfer (see Fig.~\ref{fig:teaser}).

Specifically, LIA-X is designed as a self-supervised autoencoder that does not rely on any explicit structure representations. Inspired by sparse dictionary coding~\cite{sparsecoding}, we incorporates a novel \textit{Sparse Motion Dictionary} - a set of motion vectors in autoencoder to capture the underlying motion distribution. The sparsity constraint encourages the model to use a minimal set of these motion vectors to reconstruct the training images, endowing the motion vectors with enhanced interpretability compared to the dense motion dictionary proposed in prior work~\cite{lia-pami, wang2022latent}. The animation process is then modeled as a linear navigation of these motion codes to produce optical-flow fields for warping the source portrait. Interestingly, we found that these interpretable motion vectors can also be directly leveraged at the inference stage to manipulate source portrait, supporting fine-grained edits of facial attributes (e.g., eyes and mouth) as well as 3D-aware transformation (e.g., yaw, pitch and roll) in both image and video domains.

In addition, we have analyzed the scalability of proposed LIA-X framework and successfully train large-scale models with up to 1 billion parameters using a diverse mixture of public and internal talking head datasets. Experimental results demonstrate that scaling up the LIA-X significantly improves performance across several benchmarks. Given the faster inference speed of autoencoder-based models compared to diffusion-based approaches, we believe the scalable and interpretable design of LIA-X can serve as a valuable complement to current state-of-the-art generative models, enabling efficient and highly controllable video generation. In summary, the key contributions of this work are:
\begin{itemize}
    \item We propose a novel portrait animator, LIA-X, an autoencoder that incorporates an interpretable Sparse Motion Dictionary to enable controllable portrait animation via an \textit{`edit-warp-render'} strategy.
    \item We analyze the scalability of the LIA-X and demonstrate that our design can be scaled up to achieve superior performance.
    \item Extensive experiments show that LIA-X outperforms state-of-the-art methods across several datasets, while also supporting a various of applications such as image and video editing, and 3D-aware portrait manipulation.
\end{itemize}

%% file: sec/2_related_work.tex
\begin{figure*}[!t]
    \centering
    \includegraphics[width=1.0\linewidth]{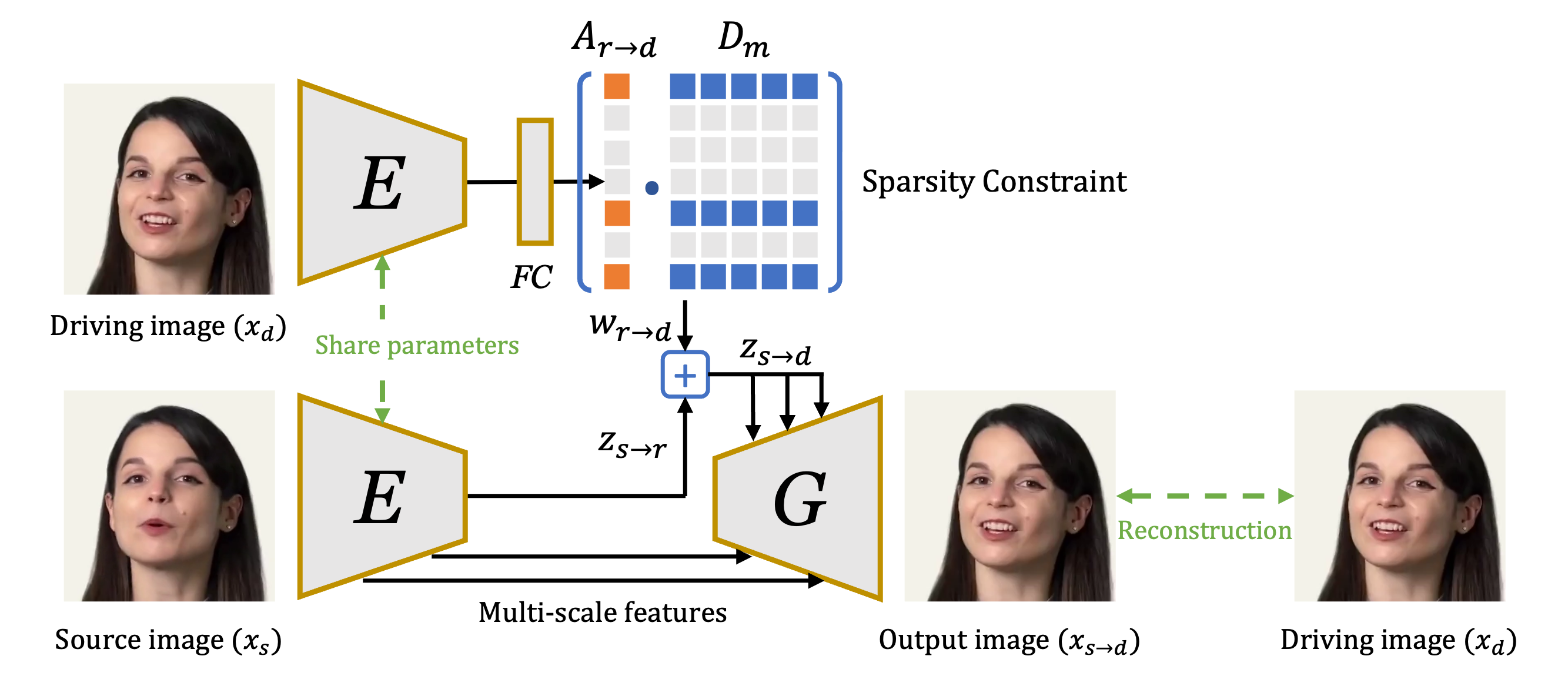}
    \caption{\textbf{Overview of LIA-X.} LIA-X consists of an encoder $E$, a generator $G$ that includes an optical-flow generator $G_f$ and a rendering network $G_r$. LIA-X is trained using a self-supervised learning strategy. To obtain an interpretable motion dictionary $D_m$, a sparsity constraint is incorporated into the training objective, enrouraging the network to use a minimal number of vectors in $D_m$ to reconstruct each driving image.}
    \label{fig:3dvideo}
\end{figure*}

\section{Related Work}
\label{sec:related_work}

Portrait animation has seen significant advancements in recent years, driven by the remarkable progress in deep generative models for video generation~\cite{vondrick2016generating,saito2017temporal,tulyakov2017mocogan,wang:hal-02368319,wang2020g3an,wang:tel-03551913,wang2021inmodegan,clark2019adversarial, brooks2022generating,digan,stylegan-v,mocoganhd,leo,seine,videocrafter2,guo2023animatediff,makeavideo,imagenvideo,latte,lavie,videoLDM,snapvideo,videogpt,sora,show-1}. Previous methods have explored learning structure representations either based on conditional generation approaches~\cite{chan2019everybody,wang2018vid2vid,zakharov2019few,wang2019fewshotvid2vid,transmomo2020,thinplate} relying on off-the-shelf extractors, or self-supervised learning strategies to learn representations such as 2D/3D keypoints~\cite{fomm, facevid2vid, liveportrait, thinplate}, motion regions~\cite{siarohin2021motion}, and depth maps~\cite{dagan} in an end-to-end manner. Self-attention mechanism~\cite{gong2023toontalker} has also been studied to improve cross-identity generation quality. 

More recent techniques~\cite{lia-pami, wang2022latent} have proposed to model motion transfer as a linear navigation of learned motion codes, which has proven effective in implicitly capturing 2D and 3D representations to capture complex facial dynamics. Diffusion-based models~\cite{x-portrait} have also been shown to be effective for portrait animation, due to their powerful generalization abilities pretrained on large-scale datasets. 

However, one key limitation still exists across all the existing methods, they lack an effective mechanism to align the source portrait with the initial driving frame in terms of head poses and facial expressions. This makes their performance drop dramatically when there are large variations between the source and driving data, further limiting their usability in more general real-world scenarios. Different from previous approaches, LIA-X incorporates an interpretable motion dictionary, which allows users to edit source portrait before animation in a controllable manner. With this unique capacity, we found LIA-X can be further used in several tasks such as portrait animation, image and video editing, and 3D-aware portrait manipulation.

%% file: sec/3_method.tex
\section{Preliminary}
Latent Image Animator (LIA)~\cite{lia-pami,wang2022latent} is designed as an autoencoder consisting of an encoder \textit{E}, an optical-flow generator \textit{$G_f$} and a rendering network \textit{$G_r$} aiming to transfer motion of a talking head to a still portrait via self-supervised learning.  LIA models motion as linear navigation of motion code in latent space, and follows the \textit{`warp-render'} strategy to generate optical-flow fields via $G_f$ and render the animated result via $G_r$.

\textbf{Linear Navigation.} LIA models motion transferring as learning transformations from source to driving image $x_{s}\rightarrow x_{d}$. It proved that for any given image, there exists an `implicit reference image' $x_{r}$, and the transformation can be modeled as $x_{s}\rightarrow x_{r}\rightarrow x_{d}$ in an implicit manner. The transformation is modeled as motion code $z_{s\rightarrow d}$ in latent space, and with the help of the reference space, it can be represented as a linear navigation from $z_{s\rightarrow r}$ along a path $w_{r\rightarrow d}$, denoted as
\begin{equation}
    z_{s\rightarrow d}=z_{s\rightarrow r}+w_{r\rightarrow d},
\end{equation}
where $z_{s\rightarrow r}$ indicates the transformation from source image to reference image, and is obtained via $E(x_s)=z_{s\rightarrow r}$. To learn $w_{r\rightarrow d}$, a learnable motion dictionary $D_m$ consisting of a set of orthogonal motion vectors $\{\mathbf{d_1},...,\mathbf{d_M}\}$ is proposed. Any linear path in the latent space can be represented as
\begin{equation}
w_{r\rightarrow d}= \sum^{M}_{i=1}a_{i}\mathbf{d_{i}},
\end{equation}
where $\mathcal{A}_{r\rightarrow d}=\{a_1,...,a_M\}$ indicates the magnitude for each motion vector, and is obtained via $\mathcal{FC}(E(x_d))=\mathcal{A}_{r\rightarrow d}$. The linear navigation from any $x_s$ to $x_d$ can be represented as
\begin{equation}
z_{s\rightarrow d}=z_{s\rightarrow r}+\sum^{M}_{i=1}a_{i}\mathbf{d_{i}}.
\label{eq:linear_navigation}
\end{equation}

\textbf{Image Animation.} Once $z_{s\rightarrow d}$ is obtained, optical flows are generated via $G_f(z_{s\rightarrow d})=\phi_{s\rightarrow d}$. The source image $x_s$ is warped based on $\phi$ and the warped features are rendered through $G_r$ to produce the generated image
\begin{equation}
x_{s\rightarrow d}=G_r(\mathcal{T}(\phi_{s\rightarrow d},x_s)),
\label{eq:linear_navigation}
\end{equation}
where $\mathcal{T}$ indicates the warping operation. In practice, multi-scale optical flows are generated to warp multi-scale feature maps of $x_s$.

\textbf{Learning.} The objective of self-supervised learning consists of three parts, an $L_1$ reconstruction loss, a VGG-based perceptual loss and an adversarial loss between $x_{s\rightarrow d}$ and $x_d$
\begin{equation}
\begin{split}
\mathcal{L}(x_{s\rightarrow d}, x_d)=\mathcal{L}_{recon}(x_{s\rightarrow d}, x_d) &+ \lambda\mathcal{L}_{vgg}(x_{s\rightarrow d}, x_d) \\ 
&+ \mathcal{L}_{adv}(x_{s\rightarrow d}).
\end{split}
\end{equation}

\textbf{Inference.} LIA proposed a unified formulation for self-reenactment and cross-reenactment at the inference stage
\begin{equation}
\begin{split}
z_{s\rightarrow t} = (z_{s\rightarrow r} + w_{r\rightarrow s}) + (w_{r\rightarrow t} - w_{r\rightarrow 1}),\;t\in \{1,...,T\},
\end{split}
\label{eq:inference}
\end{equation}

For self-reenactment, where $w_{r\rightarrow s} = w_{r\rightarrow 1}$, Eq.~\ref{eq:inference} can be simplified as 
\begin{equation}
    z_{s\rightarrow t}=z_{s\rightarrow r} + w_{r\rightarrow t},\;t\in \{1,...,T\},
\label{eq:selfreenactment}
\end{equation}
which is the same as the training stage.

For cross-reenactment, where $w_{r\rightarrow s} \ne w_{r\rightarrow 1}$, Eq.~\ref{eq:inference} can be rewritten as
\begin{equation}
\begin{split}
z_{s\rightarrow t} = \underbrace{z_{s\rightarrow s}}_{\text{reconstruction}} + \underbrace{(w_{r\rightarrow t} - w_{r\rightarrow 1})}_{\text{motion difference}},\;t\in \{1,...,T\},
\end{split}
\label{eq:crossreenactment}
\end{equation}
where the animation process is modeled as \textit{linearly navigating the source image along the driving motion direction in the latent space}. 


\section{Methodology of LIA-X}
In this section, we proceed to introduce our proposed LIA-X, including the general model architecture, the Sparse Motion Dictionary, as well as the interpretable and controllable animation capabilities.

\subsection{Architecture}
LIA-X follows the general architecture design of LIA, containing an encoder $E$, an optical-flow generator $G_f$, and a rendering network $G_r$. However, deviating from previous methods~\cite{fomm, siarohin2021motion, thinplate} that rely on simple and small-scale networks, we design the LIA-X architecture to be more scalable by incorporating advanced residual blocks inspired by StyleGAN-T~\cite{stylegan-t} in both the encoder and generators.

\subsection{Sparse Motion Dictionary}
While the original motion dictionary in LIA contains certain semantic meanings, we observed that it is difficult to disentangle and independently control individual factors such as mouth movements, eyebrow expressions, etc. Multiple semantics are entangled and hard to leverage for controllable image and video editing tasks.

Inspired by \textit{sparse dictionary coding}, we propose a Sparse Motion Dictionary to improve the interpretability of the motion representations. We found that this simple improvement is extremely effective in enhancing the interpretability of the motion vectors, which in turn enables more controllable animation.

Specifically, we introduce a sparse penalty $S(\cdot)$ on the motion coefficients $\mathcal{A}_{r\rightarrow d}$ to encourage the model to use a minimal number of motion vectors to reconstruct the images during self-supervised training. The overall learning objective of LIA-X is then:


\begin{equation}
\begin{split}
\mathcal{L}(x_{s\rightarrow d}, x_d)=&\mathcal{L}_{recon}(x_{s\rightarrow d}, x_d) + \lambda_1\mathcal{L}_{vgg}(x_{s\rightarrow d}, x_d) \\ 
&+ \mathcal{L}_{adv}(x_{s\rightarrow d}) + \lambda_2 S(\mathcal{A}_{r\rightarrow d}),
\end{split}
\end{equation}
where $\lambda_1$ and $\lambda_2$ are coefficients to balance the losses, and we implement $S(\cdot)$ as the $L_1$ norm.

\subsection{Controllable Inference}
While Eq.~\ref{eq:crossreenactment} enables successful motion transfer from the driving video to the source portrait, the high-quality generation relies on a strong requirement - \textit{the source and initial driving frames should have similar head poses and facial expressions}. Ideally, portraits with frontal pose and neutral expression usually result in the best performance. However, in various real-world applications, this requirement cannot always be satisfied. With the interpretable Sparse Motion Dictionary, LIA-X is able to address this issue by using an \textit{"edit-warp-render"} strategy.

Users are first allowed to edit the source portrait using the corresponding motion vectors, aligning the head pose and facial expression to be as similar as possible to the driving image. Then, the animation process can be formulated as:
\begin{equation}
z_{s\rightarrow t} = \underbrace{z_{s\rightarrow \mathcal{E}(s)}}_{\text{editing}} + \underbrace{(w_{r\rightarrow t} - w_{r\rightarrow 1})}_{\text{motion difference}},\;t\in \{1,...,T\},
\label{eq:relative}
\end{equation}
where $\mathcal{E}(\cdot)$ indicates the editing operation on the source image. This allows LIA-X to better handle large variations between the source and driving data, leading to higher-quality and more controllable portrait animation.



%% file: sec/4_experiments.tex
\section{Experiments}
\label{sec:exp}

\begin{figure*}[th!]
    \centering
    \includegraphics[width=\textwidth]{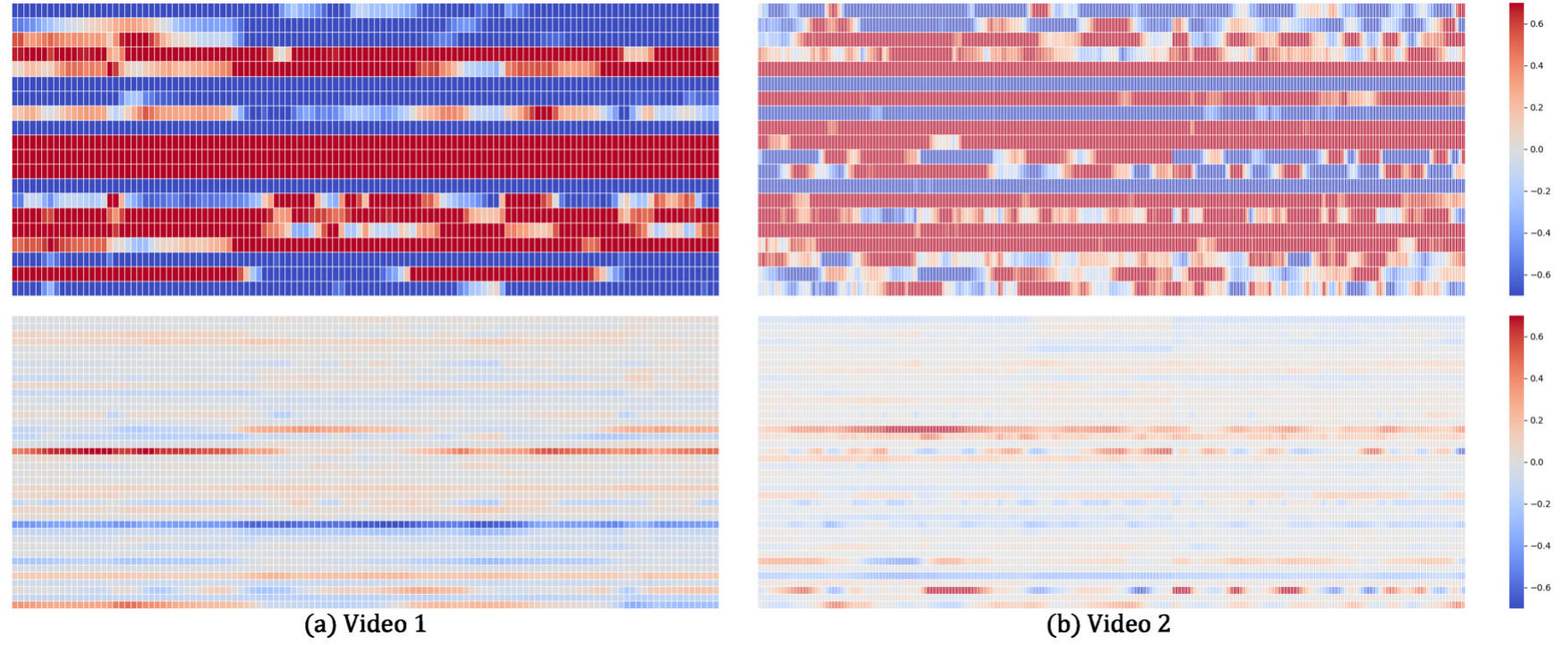}

    \caption{\textbf{Sparsity Analysis.} We show a comparison of the $\mathcal{A}_{r\rightarrow s}$ activations of two videos between two models - one trained without sparse motion dictionary (top), and the other with a sparse motion dictionary (down). It can be clearly observed that when learned without a sparsity constraint, the model reconstructs each frame by activating nearly all the motion vectors. In contrast, the model trained with a sparsity constraint selects only a few vectors to be activated for each reconstruction.}
    \label{fig:sparsity}
\end{figure*}

\begin{figure} 
\captionsetup[subfigure]{labelformat=empty}
\centering
\begin{subfigure}[t]{\linewidth}
\includegraphics[width=1.0\textwidth]{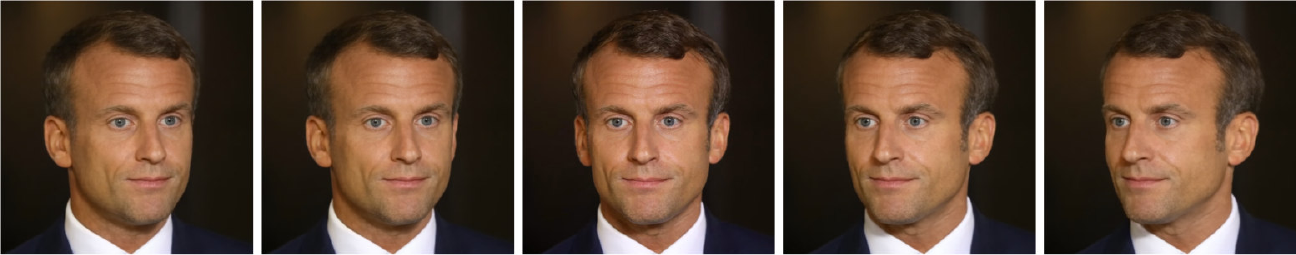}
\caption{\footnotesize{(a) Yaw}}
\end{subfigure}
\begin{subfigure}[t]{\linewidth}
\includegraphics[width=1.0\linewidth]{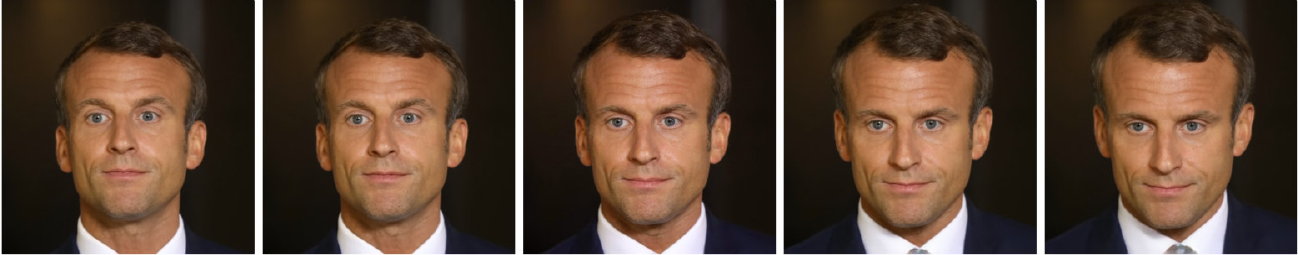}
\caption{\footnotesize{(b) Pitch}}
\end{subfigure} %
\begin{subfigure}[t]{\linewidth}    
\includegraphics[width=1.0\linewidth]{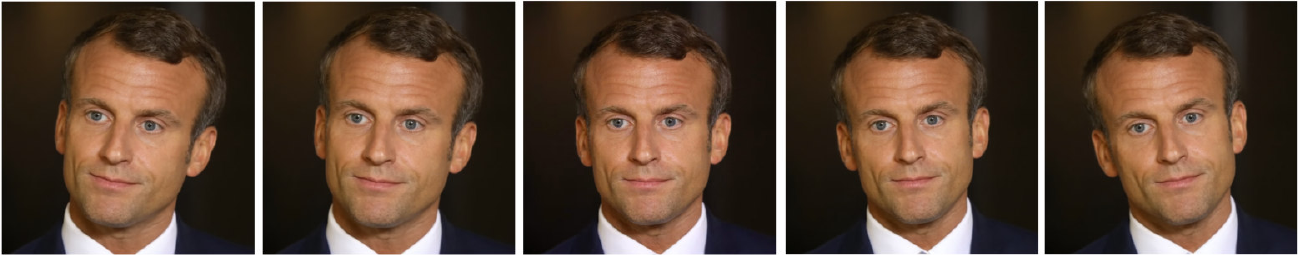}
\caption{\footnotesize{(c) Roll}}   
\end{subfigure} 
\caption{\textbf{3D-aware Portrait Manipulation.} We illustrate 3D-aware manipulation capabilities for a single identity. By manipulating corresponding motion vectors, LIA-X can successfully conduct (a) \textit{yaw}, (b) \textit{pitch} and (c) \textit{roll} adjustments, without relying on any additional 3D representations.}
\label{fig:3d-aware}
\end{figure}

\begin{figure}[!t]
\captionsetup[subfigure]{labelformat=empty}
\centering
\begin{subfigure}[t]{\linewidth}
\includegraphics[width=1.0\textwidth]{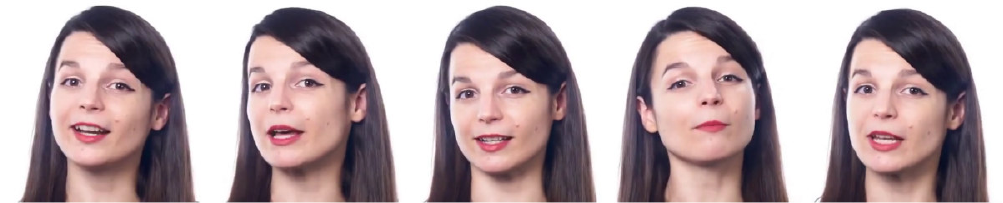}
\caption{\footnotesize{(a) Original video}}
\end{subfigure}
\begin{subfigure}[t]{\linewidth}
\includegraphics[width=1.0\linewidth]{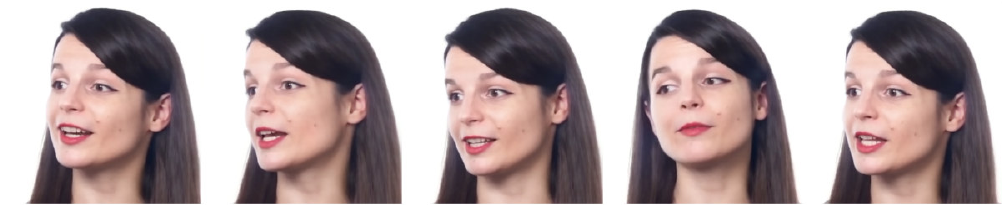}
\caption{\footnotesize{(b) Left rotation}}
\end{subfigure} %
\begin{subfigure}[t]{\linewidth}    
\includegraphics[width=1.0\linewidth]{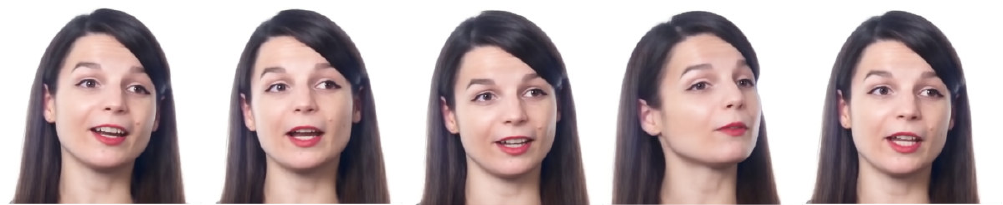}
\caption{\footnotesize{(c) Right Rotation}}   
\end{subfigure} 
\caption{\textbf{3D-aware Video Manipulation.} We demonstrate the use of learned 3D-aware semantics to manipulate a real-world video. The original video (a) has been successfully rotated to left (b) and right (c) respectively, without using any explicit 3D representation, and the identity of the original video subject has also been well-preserved throughout these manipulations.}
\label{fig:3dvideo}
\end{figure}

In this section, we will first describe the experimental setup, including implementation details and dataset information. Then, we will present a qualitative analysis on the sparsity, interpretability, and controllability of the proposed LIA-X framework. We will also show comparative results between LIA-X and state-of-the-art methods such as FOMM, TPS, DaGAN, LIA, MCNet, X-Portrait, and LivePortrait, on the task of portrait animation. Finally, we will quantitatively evaluate LIA-X and compare it with the prior work on two important tasks, \textit{self-reenactment} and \textit{cross-reenactment}.


\begin{figure*}[!t]
    \centering
    \includegraphics[width=1.0\linewidth]{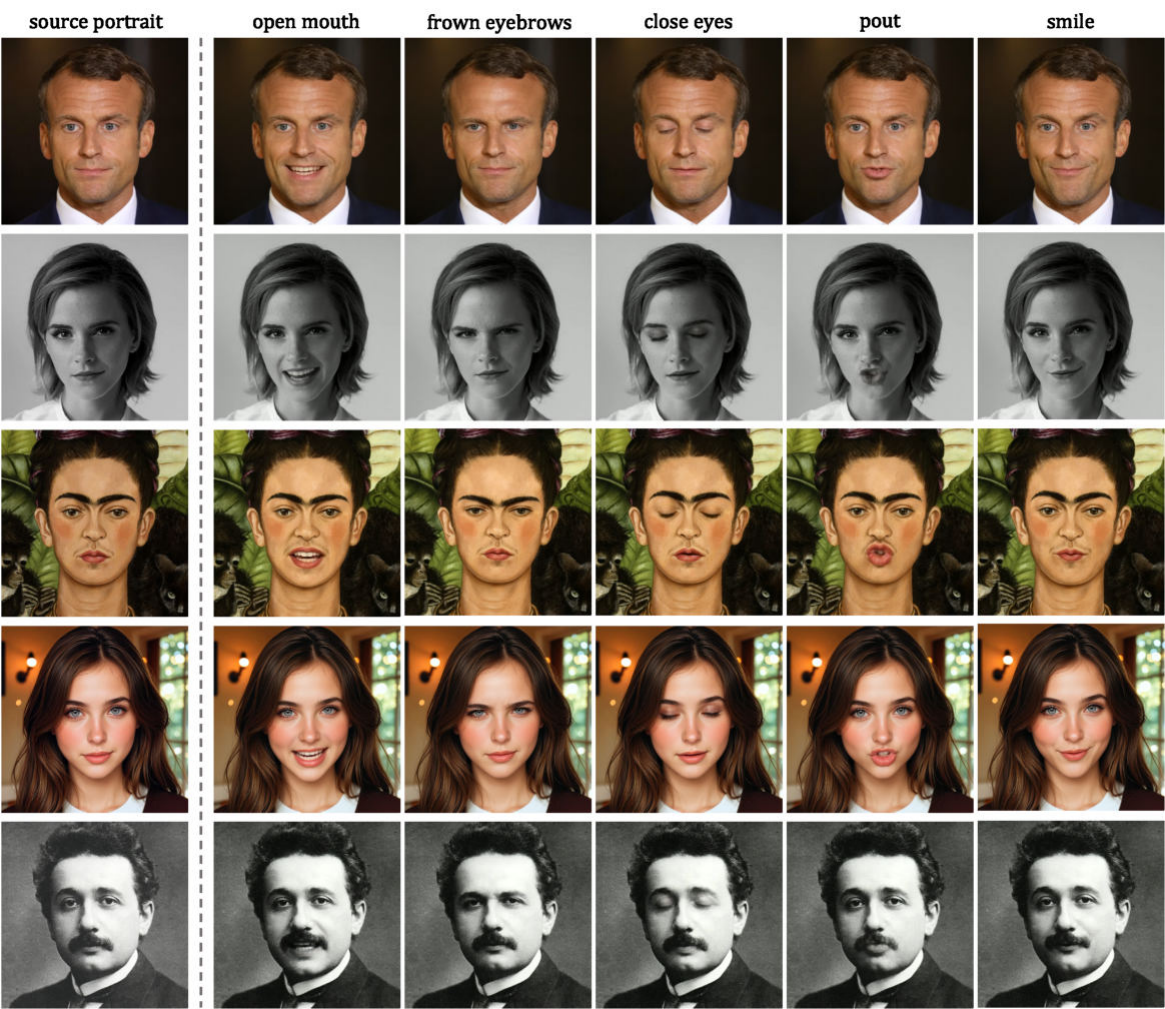}
    \caption{\textbf{Image Editing.} We show image editing capacities of LIA-X. Fine-grained semantic attributes, such as \textit{open/close mouth}, \textit{frown/raise eyebrows}, \textit{open/close eyes}, \textit{pout}, and \textit{smile} can be successfully controlled by manipulating the corresponding motion vectors.}
    \label{fig:image-edit}
\end{figure*}

\textbf{Implementation details.} We build LIA-X based on the implementation of the original LIA~\cite{lia-pami,wang2022latent}. To scale the model to larger sizes and prevent training instability, we designed novel residual blocks for both the optical flow generator $G_f$ and the rendering network $G_r$. Our scaling strategy focuses on increasing the number of channels, the depth of residual blocks, as well as using a larger motion dictionary compared to the original LIA. With these architectural enhancements, the largest model size of LIA-X reaches around 1 billion parameters. We train the entire LIA-X model using 8 A100 GPUs, and apply gradient accumulation to increase the effective batch size when training the larger-scale models.


\textbf{Datasets.} To scale the training data, we mix 4 publicly available datasets, i.e., VoxCelebHQ~\cite{lia-pami}, TalkingHead-1KH~\cite{facevid2vid}, HDTF~\cite{zhang2021flow} and MEAD~\cite{kaisiyuan2020mead}. Additionally, we include 1 internally collected dataset. In total, our training dataset contains 0.5 million talking head sequences, comprising around 94 million frames and 55,000 different identities. Experiments show that this dataset scaling strategy enables LIA-X to achieve outstanding performance in generalizing to unseen portrait images.


\begin{figure*}[!th]
    \centering
    \includegraphics[width=1.0\linewidth]{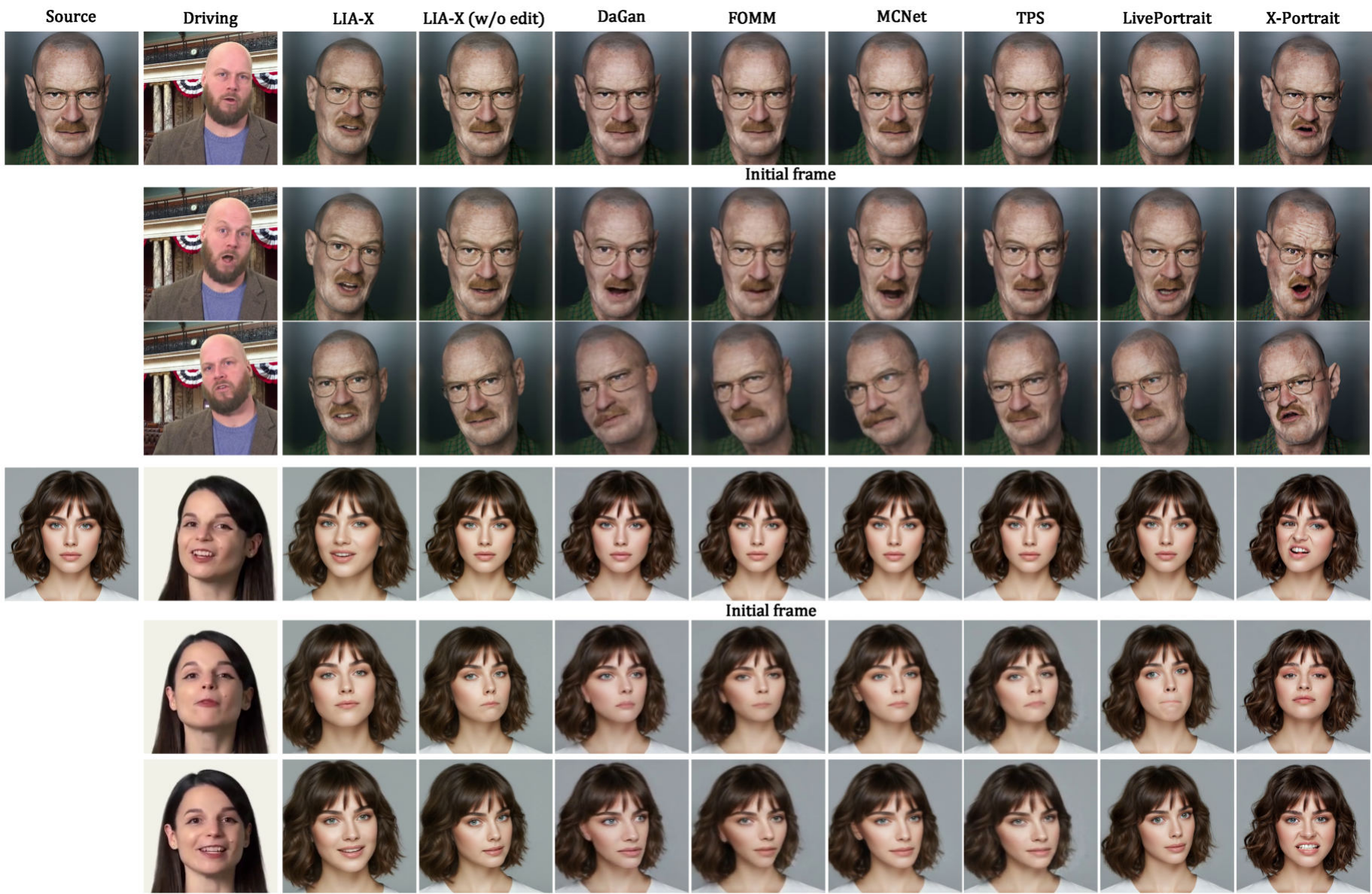}
    \caption{\textbf{Qualitative Comparison on Cross-reenactment.} LIA-X's ability to edit the source portrait using interpretable motion vectors before animation allows it to better adjust for initial misalignments with the driving frame. As shown, this editing capability enables LIA-X to significantly outperform other approaches, especially when there are large variations in head pose and facial expression between the source and driving data.}
    \label{fig:cross-id}
\end{figure*}

\subsection{Sparsity Analysis}
To demonstrate the effectiveness of our proposed Sparse Motion Dictionary, we visualized the motion coefficient vectors $\mathcal{A}_{r\rightarrow s}$ for two different videos, as shown in Fig.~\ref{fig:sparsity}. We compare the results of models with and without using the sparsity constraint in the motion dictionary. 

It can be clearly observed that the model without the sparse motion dictionary activates almost all the motion vectors, indicating a lack of selectivity for different input data. The semantics are entangled within each motion vector. In contrast, for the model with the Sparse Motion Dictionary, sparsity can be clearly observed in $\mathcal{A}_{r\rightarrow s}$ - only a few vectors are active, while the contributions of the others can be largely omitted. These results clearly prove the effectiveness of the constraint in improving the sparsity of the motion dictionary.



\subsection{Interpretability and Controllability Analysis}
While we have proven the ability to obtain a Sparse Motion Dictionary, it is crucial to understand whether each vector in the dictionary is interpretable and controllable. To investigate the semantic meanings of the motion vectors, we linearly manipulate the vector $d_i$ using:
\begin{equation}
\begin{split}
z_{s\rightarrow \mathcal{E}(s)} = z_{s\rightarrow s} + a_{i}\mathbf{d_{i}},
\end{split}
\label{eq:manipulation}
\end{equation}
where we set $a_i$ as a small perturbation ranging from -0.5 to 0.5 with a step of 0.1. Surprisingly, we found that the semantics are well-disentangled in the motion dictionary and can be easily manipulated. Almost all the activated vectors correspond to human-understandable meanings. Fig.~\ref{fig:3d-aware} shows examples of LIA-X performing 3D-aware manipulations such as\textit{yaw}, \textit{pitch} and \textit{roll}, without relying on any explicit 3D representation during training or inference.


\textbf{Image Editing.} We further utilize Eq.~\ref{eq:manipulation} to manipulate more motion vectors and demonstrate the results in Fig.~\ref{fig:image-edit}. Besides the 3D-aware semantics, LIA-X can control various fine-grained attributes like `open/close mouth', `frown/raise eyebrows', 'open/close eyes', `pout', `smile', etc. All these interpretable vectors are automatically disentangled through self-supervised learning. With such a powerful editing capability, LIA-X can be readily used as a portrait editing tool, allowing users to linearly combine different semantic operations to achieve complex editing tasks.

\begin{table*}[!th]
    \centering
    \scalebox{0.95}{
    \begin{tabular}{cccccccccccc}
    \hline
        \multirow{3}{*}{Method} 
        & \multicolumn{5}{c}{\textbf{VoxCelebHQ}} & \multicolumn{5}{c}{\textbf{TalkingHead-1KH}} \\
        \cline{2-6} \cline{8-12}
        & \textbf{L1}~$\downarrow$ & \textbf{LPIPS}~$\downarrow$ & \textbf{SSIM}~$\uparrow$  & \textbf{PSNR}~$\uparrow$ & \textbf{FID}~$\downarrow$ && \textbf{L1}~$\downarrow$ & \textbf{LPIPS}~$\downarrow$ & \textbf{SSIM}~$\uparrow$  & \textbf{PSNR}~$\uparrow$ & \textbf{FID}~$\downarrow$ \\
    \cline{2-12}
    & \multicolumn{11}{c}{$256\times 256$ resolution} \\
    \hline
        FOMM~\cite{fomm} & 0.046 & 0.27 & 0.66 & 22.40 & 12.67 && 0.040 & 0.100 & 0.72 & 23.31 & 30.39 \\
        DaGAN~\cite{dagan} & 0.044 & 0.110 & 0.69 & 23.04 & \textbf{9.13} &&0.036 & 0.088 & 0.77 & 24.95 & 25.50 \\
        TPS~\cite{thinplate} & 0.043 & 0.112 & 0.70 & 23.24 & 10.82 && 0.037 & 0.089 & 0.77 & 24.56 & 28.05 \\
        MCNet~\cite{mcnet} & 0.040 & 0.176 & 0.72 & 23.73 & 18.63 && \textbf{0.030} & 0.097 & \textbf{0.79} & 25.70 & 28.06 \\
    \hline
        LIA-X & \textbf{0.036} & \textbf{0.095} & \textbf{0.73} & \textbf{24.82} & 10.74 && 0.035 & \textbf{0.086} & 0.78 & \textbf{26.26} & \textbf{24.71} \\
    \hline
    \hline
    & \multicolumn{11}{c}{$512\times 512$ resolution} \\
    \hline
        X-Portrait~\cite{x-portrait} & 0.110 & 0.302 & 0.56 & 16.99 & 19.92 && 0.058 & 0.134 & 0.63 & 19.46 & 41.19 \\
        LivePortrait~\cite{liveportrait} & 0.087 & 0.264 & 0.67 & 17.45 & 12.90 && 0.052 & 0.120 & 0.73 & 20.26 & 39.98 \\
        LIA~\cite{lia-pami} & 0.052 & 0.211 & 0.68 & 22.14 & 21.86 && 0.049 & 0.165 & 0.72 & 23.37 & 44.64 \\
    \hline
        LIA-X & \textbf{0.040} & \textbf{0.160} & \textbf{0.75} & \textbf{24.39} & \textbf{12.50} && \textbf{0.035} & \textbf{0.115} & \textbf{0.80} & \textbf{26.07} & \textbf{38.93} \\
    \hline
    \end{tabular}}
    \caption{\textbf{Quantitative Evaluation for Self-Reenactment.} We compare the performance of LIA-X against state-of-the-art methods on two different resolutions across two datasets.}
    \label{tab:self}
\end{table*}

\textbf{Video Editing.} Once we can edit a single portrait, extending the editing to the video level becomes feasible. Given a real-world talking head video, we first apply Eq.~\ref{eq:manipulation} on the initial frame to manipulate the target semantics. Then, we use Eq.~\ref{eq:relative} to transfer the motion to the edited first frame. Fig.~\ref{fig:3dvideo} demonstrates the results of using LIA-X to rotate the portrait in a video, where the identity is well-preserved while the head pose is seamlessly changed.

\begin{table}[!t]
    \centering
    \scalebox{0.95}{
    \begin{tabular}{ccc}
    \hline
        Method & \textbf{ID Similarity}~$\downarrow$ & \textbf{Image Quality}~$\uparrow$ \\
    \hline
        FOMM~\cite{fomm} & 0.262 & 37.08 \\
        DaGAN~\cite{dagan} & 0.272 & 39.30 \\
        TPS~\cite{thinplate} & 0.216 & 38.27 \\
        MCNet~\cite{mcnet} & 0.252 & 37.88 \\
        X-Portrait~\cite{x-portrait} & 0.217 & 55.41 \\
        LivePortrait~\cite{liveportrait} & 0.243 & 51.41\\
    \hline
        LIA-X & \textbf{0.206} & \textbf{58.74}  \\
    \hline
    \end{tabular}}
    \caption{\textbf{Quantitative Evaluation for Cross-Reenactment.} We compare the performance of LIA-X against state-of-the-art methods on the constructed dataset for cross-reenactment scenarios.}
    \label{tab:cross}
\end{table}

\textbf{Portrait Animation.} We also qualitatively compare LIA-X with state-of-the-art methods on the challenging task of cross-identity portrait animation, as shown in Fig.~\ref{fig:cross-id}. The visualizations clearly showcase the effectiveness of LIA-X's editing stage, which can significantly improve the generated results by aligning the source portrait with the initial driving frame. This editing capability allows for a more controllable animation process compared to previous techniques. In contrast, other state-of-the-art methods struggle when the source and driving data have large variations in head poses, facial expressions, and identities. Their performance degrades dramatically in such cross-identity scenarios, whereas LIA-X maintains high-quality and controllable animation results by leveraging its interpretable motion representations and the tailored "edit-warp-render" pipeline. This comparison highlights a key advantage of our proposed framework - its ability to handle large discrepancies between the source and driving data through the initial editing stage, before seamlessly applying the motion transfer.

\begin{table}[!t]
    \centering
    \scalebox{0.95}{\begin{tabular}{ccccc}
    \hline
        Model & \textbf{L1}~$\downarrow$ & \textbf{LPIPS}~$\downarrow$ & \textbf{SSIM}~$\uparrow$ & \textbf{PSNR}~$\uparrow$ \\
    \hline
        Base (0.05B) & 0.043 & 0.171 & 0.72 & 23.62\\
        Middle (0.3B) & \textbf{0.040} & \textbf{0.16} & 0.74 & 24.31 \\
        Large (0.9B) & \textbf{0.040} & \textbf{0.16} & \textbf{0.75} & \textbf{24.39} \\
    \hline
    \end{tabular}}
    \caption{\textbf{Scalability analysis on VoxCelebHQ.} }
    \label{tab:ablation1}
\end{table}

\begin{table}[!t]
    \centering
    \scalebox{0.95}{\begin{tabular}{ccccc}
    \hline
        Model & \textbf{L1}~$\downarrow$ & \textbf{LPIPS}~$\downarrow$ & \textbf{SSIM}~$\uparrow$ & \textbf{PSNR}~$\uparrow$ \\
    \hline
        Base (0.05B) & 0.042 & 0.13 & 0.77 & 24.98 \\
        Middle (0.3B) & \textbf{0.035} & \textbf{0.113} & 0.79 & 25.84 \\
        Large (0.9B) & \textbf{0.035} & 0.115 & \textbf{0.80} & \textbf{26.07} \\
    \hline
    \end{tabular}}
    \caption{\textbf{Scalability analysis on TalkingHead-1KH.} }
    \label{tab:ablation2}
\end{table}

\subsection{Quantitative Evaluation}
We quantitatively compare LIA-X with state-of-the-art methods on two important tasks, \textit{self-reenactment} and \textit{cross-reenactment}. To ensure a fair comparison, we train LIA-X at two different resolutions, $256\times256$ and $512\times512$, and compare with the corresponding methods.

\textbf{Self-Reenactment.} We evaluate our method on the validation sets of VoxCelebHQ~\cite{lia-pami} and TalkingHead-1KH~\cite{facevid2vid}, which contain 483 and 25 videos respectively. We reconstruct each video sequence by using the first frame as the source image and the entire video as the driving data. We report results using several metrics, L1, LPIPS, SSIM, PSNR, and FID. As shown in Tab.~\ref{tab:self}, on both high and low resolutions, LIA-X outperforms both GAN-based and diffusion-based state-of-the-art methods across all the metrics, demonstrating the effectiveness of our proposed architecture design and training strategy.

\textbf{Cross-Reenactment.} To construct the validation set for cross-reenactment, we select 70 videos from the HDTF dataset~\cite{zhang2021flow} as the driving data, and for each video, we randomly select 2 images from the AAHQ dataset~\cite{liu2021blendgan} as the source data, resulting in a total of 140 videos. Since there is no ground truth data for this task, we evaluate the results using two metrics, \textit{Identity Similarity} and \textit{Image Quality}. Identity Similarity measures the average embedding difference between each generated frame and the source portrait, while Image Quality is computed following ~\cite{Su_2020_CVPR} to indicate the generated image quality. As shown in Tab.~\ref{tab:cross}, LIA-X outperforms other methods in both metrics, proving its ability to naturally transfer motion while preserving the original identities.

\subsection{Ablation study on scalability}
To verify the effectiveness of scalability, we conduct an ablation study on the model size. We train three variations of LIA-X with 0.05B, 0.3B, and 0.9B parameters, respectively, while keeping the same training configuration. The three models follow the same architectural design and only differ in the number of residual blocks, channel numbers, and motion dictionary size. The results in Tab.~\ref{tab:ablation1} and ~\ref{tab:ablation2} clearly demonstrate that scaling is effective for improving model performance. However, we note that the improvement from 0.3 billion to 0.9 billion parameters becomes relatively minor. We hypothesize that this could be due to the current dataset size not being large enough to fully support training such a large-scale model. To further push the limits of LIA-X's scalability and performance, we believe the involvement of even larger and more diverse training datasets would be necessary.

%% file: sec/5_limitation.tex
\section{Limitations and future work}
\label{limitation}

While our proposed method has achieved promising performance, there are still limitations need to be tackled in future work. Firstly, current model can only be used on fixed resolution, designing novel technique to allow dynamic resolution may further improve the performance. Secondly, our model follows the convolutional network architecture which may have limitations to further scale up. DiT-based~\cite{Peebles2022DiT} architecture should be explored in future work for scalability.

%% file: sec/6_conclusion.tex
\section{Conclusion}
\label{sec:conclusion}

In this work, we have introduced LIA-X, a novel portrait animator that incorporates an interpretable Sparse Motion Dictionary. This enables LIA-X to support a highly controllable `edit-warp-render' animation strategy. Furthermore, we have analyzed the scalability of the LIA-X architecture, demonstrating its ability to achieve outstanding results when scaling up to larger model sizes, up to 1 billion parameters. Extensive evaluations show that LIA-X outperforms SOTA methods across several benchmarks, while also supporting diverse applications like fine-grained image/video editing and high-quality portrait animation. We envision our technique serve as a valuable complement to current generative approaches and provide a novel way for interpretable video generation.

%% file: main.bbl
\begin{thebibliography}{58}
\providecommand{\natexlab}[1]{#1}
\providecommand{\url}[1]{\texttt{#1}}
\expandafter\ifx\csname urlstyle\endcsname\relax
  \providecommand{\doi}[1]{doi: #1}\else
  \providecommand{\doi}{doi: \begingroup \urlstyle{rm}\Url}\fi

\bibitem[Blattmann et~al.(2023)Blattmann, Rombach, Ling, Dockhorn, Kim, Fidler, and Kreis]{videoLDM}
Andreas Blattmann, Robin Rombach, Huan Ling, Tim Dockhorn, Seung~Wook Kim, Sanja Fidler, and Karsten Kreis.
\newblock Align your latents: High-resolution video synthesis with latent diffusion models.
\newblock In \emph{CVPR}, 2023.

\bibitem[Brooks et~al.(2022)Brooks, Hellsten, Aittala, Wang, Aila, Lehtinen, Liu, Efros, and Karras]{brooks2022generating}
Tim Brooks, Janne Hellsten, Miika Aittala, Ting-Chun Wang, Timo Aila, Jaakko Lehtinen, Ming-Yu Liu, Alexei~A Efros, and Tero Karras.
\newblock Generating long videos of dynamic scenes.
\newblock 2022.

\bibitem[Brooks et~al.(2024)Brooks, Peebles, Holmes, DePue, Guo, Jing, Schnurr, Taylor, Luhman, Luhman, Ng, Wang, and Ramesh]{sora}
Tim Brooks, Bill Peebles, Connor Holmes, Will DePue, Yufei Guo, Li Jing, David Schnurr, Joe Taylor, Troy Luhman, Eric Luhman, Clarence Ng, Ricky Wang, and Aditya Ramesh.
\newblock Video generation models as world simulators.
\newblock 2024.

\bibitem[Chan et~al.(2019)Chan, Ginosar, Zhou, and Efros]{chan2019everybody}
Caroline Chan, Shiry Ginosar, Tinghui Zhou, and Alexei~A Efros.
\newblock Everybody dance now.
\newblock In \emph{ICCV}, 2019.

\bibitem[Chang et~al.(2023)Chang, Shi, Gao, Fu, Xu, Song, Yan, Yang, and Soleymani]{magicdance}
Di Chang, Yichun Shi, Quankai Gao, Jessica Fu, Hongyi Xu, Guoxian Song, Qing Yan, Xiao Yang, and Mohammad Soleymani.
\newblock Magicdance: Realistic human dance video generation with motions \& facial expressions transfer.
\newblock \emph{arXiv preprint arXiv:2311.12052}, 2023.

\bibitem[Chen et~al.(2021)Chen, Wang, Lagadec, Dantcheva, and Bremond]{Chen_2021_CVPR}
Hao Chen, Yaohui Wang, Benoit Lagadec, Antitza Dantcheva, and Francois Bremond.
\newblock Joint generative and contrastive learning for unsupervised person re-identification.
\newblock In \emph{CVPR}, 2021.

\bibitem[Chen et~al.(2024)Chen, Zhang, Cun, Xia, Wang, Weng, and Shan]{videocrafter2}
Haoxin Chen, Yong Zhang, Xiaodong Cun, Menghan Xia, Xintao Wang, Chao Weng, and Ying Shan.
\newblock Videocrafter2: Overcoming data limitations for high-quality video diffusion models.
\newblock In \emph{CVPR}, 2024.

\bibitem[Chen et~al.(2023)Chen, Wang, Zhang, Zhuang, Ma, Yu, Wang, Lin, Qiao, and Liu]{seine}
Xinyuan Chen, Yaohui Wang, Lingjun Zhang, Shaobin Zhuang, Xin Ma, Jiashuo Yu, Yali Wang, Dahua Lin, Yu Qiao, and Ziwei Liu.
\newblock Seine: Short-to-long video diffusion model for generative transition and prediction.
\newblock In \emph{ICLR}, 2023.

\bibitem[Clark et~al.(2019)Clark, Donahue, and Simonyan]{clark2019adversarial}
Aidan Clark, Jeff Donahue, and Karen Simonyan.
\newblock Adversarial video generation on complex datasets.
\newblock \emph{arXiv preprint arXiv:1907.06571}, 2019.

\bibitem[Goodfellow et~al.(2014)Goodfellow, Pouget-Abadie, Mirza, Xu, Warde-Farley, Ozair, Courville, and Bengio]{goodfellow2014generative}
Ian Goodfellow, Jean Pouget-Abadie, Mehdi Mirza, Bing Xu, David Warde-Farley, Sherjil Ozair, Aaron Courville, and Yoshua Bengio.
\newblock Generative adversarial nets.
\newblock In \emph{NIPS}, 2014.

\bibitem[Guo et~al.(2024{\natexlab{a}})Guo, Zhang, Liu, Zhong, Zhang, Wan, and Zhang]{liveportrait}
Jianzhu Guo, Dingyun Zhang, Xiaoqiang Liu, Zhizhou Zhong, Yuan Zhang, Pengfei Wan, and Di Zhang.
\newblock Liveportrait: Efficient portrait animation with stitching and retargeting control.
\newblock \emph{arXiv preprint arXiv:2407.03168}, 2024{\natexlab{a}}.

\bibitem[Guo et~al.(2024{\natexlab{b}})Guo, Yang, Rao, Liang, Wang, Qiao, Agrawala, Lin, and Dai]{guo2023animatediff}
Yuwei Guo, Ceyuan Yang, Anyi Rao, Zhengyang Liang, Yaohui Wang, Yu Qiao, Maneesh Agrawala, Dahua Lin, and Bo Dai.
\newblock Animatediff: Animate your personalized text-to-image diffusion models without specific tuning.
\newblock \emph{International Conference on Learning Representations}, 2024{\natexlab{b}}.

\bibitem[Ho et~al.(2020)Ho, Jain, and Abbeel]{ddpm}
Jonathan Ho, Ajay Jain, and Pieter Abbeel.
\newblock Denoising diffusion probabilistic models.
\newblock \emph{Advances in Neural Information Processing Systems}, 33:\penalty0 6840--6851, 2020.

\bibitem[Ho et~al.(2022)Ho, Chan, Saharia, Whang, Gao, Gritsenko, Kingma, Poole, Norouzi, Fleet, et~al.]{imagenvideo}
Jonathan Ho, William Chan, Chitwan Saharia, Jay Whang, Ruiqi Gao, Alexey Gritsenko, Diederik~P Kingma, Ben Poole, Mohammad Norouzi, David~J Fleet, et~al.
\newblock Imagen video: High definition video generation with diffusion models.
\newblock \emph{arXiv preprint arXiv:2210.02303}, 2022.

\bibitem[Hong and Xu(2023)]{mcnet}
Fa-Ting Hong and Dan Xu.
\newblock Implicit identity representation conditioned memory compensation network for talking head video generation.
\newblock In \emph{ICCV}, 2023.

\bibitem[Hong et~al.(2022)Hong, Zhang, Shen, and Xu]{dagan}
Fa-Ting Hong, Longhao Zhang, Li Shen, and Dan Xu.
\newblock Depth-aware generative adversarial network for talking head video generation.
\newblock 2022.

\bibitem[Li et~al.(2018)Li, Fang, Yang, Wang, Lu, and Yang]{li2018flow}
Yijun Li, Chen Fang, Jimei Yang, Zhaowen Wang, Xin Lu, and Ming-Hsuan Yang.
\newblock Flow-grounded spatial-temporal video prediction from still images.
\newblock In \emph{ECCV}, 2018.

\bibitem[Liu et~al.(2021)Liu, Li, Qin, Zhang, Wan, and Zheng]{liu2021blendgan}
Mingcong Liu, Qiang Li, Zekui Qin, Guoxin Zhang, Pengfei Wan, and Wen Zheng.
\newblock Blendgan: Implicitly gan blending for arbitrary stylized face generation.
\newblock In \emph{Advances in Neural Information Processing Systems}, 2021.

\bibitem[Liu et~al.(2019)Liu, Piao, Min, Luo, Ma, and Gao]{liu2019liquid}
Wen Liu, Zhixin Piao, Jie Min, Wenhan Luo, Lin Ma, and Shenghua Gao.
\newblock Liquid warping gan: A unified framework for human motion imitation, appearance transfer and novel view synthesis.
\newblock In \emph{CVPR}, 2019.

\bibitem[Ma et~al.(2025)Ma, Wang, Chen, Jia, Liu, Li, Chen, and Qiao]{latte}
Xin Ma, Yaohui Wang, Xinyuan Chen, Gengyun Jia, Ziwei Liu, Yuan-Fang Li, Cunjian Chen, and Yu Qiao.
\newblock Latte: Latent diffusion transformer for video generation.
\newblock \emph{Transactions on Machine Learning Research}, 2025.

\bibitem[Ma et~al.(2024)Ma, Liu, Wang, Pan, He, Yuan, Zeng, Cai, Shum, Liu, et~al.]{ma2024followyouremoji}
Yue Ma, Hongyu Liu, Hongfa Wang, Heng Pan, Yingqing He, Junkun Yuan, Ailing Zeng, Chengfei Cai, Heung-Yeung Shum, Wei Liu, et~al.
\newblock Follow-your-emoji: Fine-controllable and expressive freestyle portrait animation.
\newblock \emph{arXiv preprint arXiv:2406.01900}, 2024.

\bibitem[Menapace et~al.(2024)Menapace, Siarohin, Skorokhodov, Deyneka, Chen, Kag, Fang, Stoliar, Ricci, Ren, et~al.]{snapvideo}
Willi Menapace, Aliaksandr Siarohin, Ivan Skorokhodov, Ekaterina Deyneka, Tsai-Shien Chen, Anil Kag, Yuwei Fang, Aleksei Stoliar, Elisa Ricci, Jian Ren, et~al.
\newblock Snap video: Scaled spatiotemporal transformers for text-to-video synthesis.
\newblock In \emph{CVPR}, 2024.

\bibitem[Ohnishi et~al.(2018)Ohnishi, Yamamoto, Ushiku, and Harada]{ohnishi2018ftgan}
Katsunori Ohnishi, Shohei Yamamoto, Yoshitaka Ushiku, and Tatsuya Harada.
\newblock Hierarchical video generation from orthogonal information: Optical flow and texture.
\newblock In \emph{AAAI}, 2018.

\bibitem[Olshausen and Field(1996)]{sparsecoding}
Bruno~A Olshausen and David~J Field.
\newblock Emergence of simple-cell receptive field properties by learning a sparse code for natural images.
\newblock \emph{Nature}, 381\penalty0 (6583):\penalty0 607--609, 1996.

\bibitem[Peebles and Xie(2022)]{Peebles2022DiT}
William Peebles and Saining Xie.
\newblock Scalable diffusion models with transformers.
\newblock \emph{arXiv preprint arXiv:2212.09748}, 2022.

\bibitem[Saito et~al.(2017)Saito, Matsumoto, and Saito]{saito2017temporal}
Masaki Saito, Eiichi Matsumoto, and Shunta Saito.
\newblock Temporal generative adversarial nets with singular value clipping.
\newblock In \emph{ICCV}, 2017.

\bibitem[Sauer et~al.(2023)Sauer, Karras, Laine, Geiger, and Aila]{stylegan-t}
Axel Sauer, Tero Karras, Samuli Laine, Andreas Geiger, and Timo Aila.
\newblock {StyleGAN-T}: Unlocking the power of {GANs} for fast large-scale text-to-image synthesis.
\newblock 2023.

\bibitem[Siarohin et~al.(2019)Siarohin, Lathuili\`{e}re, Tulyakov, Ricci, and Sebe]{fomm}
Aliaksandr Siarohin, St\'{e}phane Lathuili\`{e}re, Sergey Tulyakov, Elisa Ricci, and Nicu Sebe.
\newblock First order motion model for image animation.
\newblock In \emph{NeurIPS}, 2019.

\bibitem[Siarohin et~al.(2021)Siarohin, Woodford, Ren, Chai, and Tulyakov]{siarohin2021motion}
Aliaksandr Siarohin, Oliver Woodford, Jian Ren, Menglei Chai, and Sergey Tulyakov.
\newblock Motion representations for articulated animation.
\newblock In \emph{CVPR}, 2021.

\bibitem[Singer et~al.(2023)Singer, Polyak, Hayes, Yin, An, Zhang, Hu, Yang, Ashual, Gafni, Parikh, Gupta, and Taigman]{makeavideo}
Uriel Singer, Adam Polyak, Thomas Hayes, Xi Yin, Jie An, Songyang Zhang, Qiyuan Hu, Harry Yang, Oron Ashual, Oran Gafni, Devi Parikh, Sonal Gupta, and Yaniv Taigman.
\newblock Make-a-video: Text-to-video generation without text-video data.
\newblock In \emph{ICLR}, 2023.

\bibitem[Skorokhodov et~al.(2022)Skorokhodov, Tulyakov, and Elhoseiny]{stylegan-v}
Ivan Skorokhodov, Sergey Tulyakov, and Mohamed Elhoseiny.
\newblock Stylegan-v: A continuous video generator with the price, image quality and perks of stylegan2.
\newblock In \emph{CVPR}, 2022.

\bibitem[Song et~al.(2021)Song, Meng, and Ermon]{ddim}
Jiaming Song, Chenlin Meng, and Stefano Ermon.
\newblock Denoising diffusion implicit models.
\newblock In \emph{International Conference on Learning Representations}, 2021.

\bibitem[Su et~al.(2020)Su, Yan, Zhu, Zhang, Ge, Sun, and Zhang]{Su_2020_CVPR}
Shaolin Su, Qingsen Yan, Yu Zhu, Cheng Zhang, Xin Ge, Jinqiu Sun, and Yanning Zhang.
\newblock Blindly assess image quality in the wild guided by a self-adaptive hyper network.
\newblock In \emph{CVPR}, 2020.

\bibitem[Tian et~al.(2021)Tian, Ren, Chai, Olszewski, Peng, Metaxas, and Tulyakov]{mocoganhd}
Yu Tian, Jian Ren, Menglei Chai, Kyle Olszewski, Xi Peng, Dimitris~N. Metaxas, and Sergey Tulyakov.
\newblock A good image generator is what you need for high-resolution video synthesis.
\newblock In \emph{ICLR}, 2021.

\bibitem[Tulyakov et~al.(2018)Tulyakov, Liu, Yang, and Kautz]{tulyakov2017mocogan}
Sergey Tulyakov, Ming-Yu Liu, Xiaodong Yang, and Jan Kautz.
\newblock {MoCoGAN}: Decomposing motion and content for video generation.
\newblock In \emph{CVPR}, 2018.

\bibitem[Vondrick et~al.(2016)Vondrick, Pirsiavash, and Torralba]{vondrick2016generating}
Carl Vondrick, Hamed Pirsiavash, and Antonio Torralba.
\newblock Generating videos with scene dynamics.
\newblock In \emph{NIPS}, 2016.

\bibitem[Wang et~al.(2020{\natexlab{a}})Wang, Wu, Song, Yang, Wu, Qian, He, Qiao, and Loy]{kaisiyuan2020mead}
Kaisiyuan Wang, Qianyi Wu, Linsen Song, Zhuoqian Yang, Wayne Wu, Chen Qian, Ran He, Yu Qiao, and Chen~Change Loy.
\newblock Mead: A large-scale audio-visual dataset for emotional talking-face generation.
\newblock In \emph{ECCV}, 2020{\natexlab{a}}.

\bibitem[Wang et~al.(2018)Wang, Liu, Zhu, Liu, Tao, Kautz, and Catanzaro]{wang2018vid2vid}
Ting-Chun Wang, Ming-Yu Liu, Jun-Yan Zhu, Guilin Liu, Andrew Tao, Jan Kautz, and Bryan Catanzaro.
\newblock Video-to-video synthesis.
\newblock In \emph{NeurIPS}, 2018.

\bibitem[Wang et~al.(2019)Wang, Liu, Tao, Liu, Kautz, and Catanzaro]{wang2019fewshotvid2vid}
Ting-Chun Wang, Ming-Yu Liu, Andrew Tao, Guilin Liu, Jan Kautz, and Bryan Catanzaro.
\newblock Few-shot video-to-video synthesis.
\newblock In \emph{NeurIPS}, 2019.

\bibitem[Wang et~al.(2021{\natexlab{a}})Wang, Mallya, and Liu]{facevid2vid}
Ting-Chun Wang, Arun Mallya, and Ming-Yu Liu.
\newblock One-shot free-view neural talking-head synthesis for video conferencing.
\newblock In \emph{Proceedings of the IEEE Conference on Computer Vision and Pattern Recognition}, 2021{\natexlab{a}}.

\bibitem[Wang(2021)]{wang:tel-03551913}
Yaohui Wang.
\newblock \emph{{Learning to Generate Human Videos}}.
\newblock Theses, {Inria - Sophia Antipolis ; Universit{\'e} Cote d'Azur}, 2021.

\bibitem[Wang et~al.(2020{\natexlab{b}})Wang, Bilinski, Bremond, and Dantcheva]{wang2020g3an}
Yaohui Wang, Piotr Bilinski, Francois Bremond, and Antitza Dantcheva.
\newblock {G3AN}: Disentangling appearance and motion for video generation.
\newblock In \emph{CVPR}, 2020{\natexlab{b}}.

\bibitem[Wang et~al.(2020{\natexlab{c}})Wang, Bilinski, Bremond, and Dantcheva]{wang:hal-02368319}
Yaohui Wang, Piotr Bilinski, Francois~F Bremond, and Antitza Dantcheva.
\newblock {ImaGINator: Conditional Spatio-Temporal GAN for Video Generation}.
\newblock In \emph{WACV}, 2020{\natexlab{c}}.

\bibitem[Wang et~al.(2021{\natexlab{b}})Wang, Bremond, and Dantcheva]{wang2021inmodegan}
Yaohui Wang, Francois Bremond, and Antitza Dantcheva.
\newblock Inmodegan: Interpretable motion decomposition generative adversarial network for video generation.
\newblock \emph{arXiv preprint arXiv:2101.03049}, 2021{\natexlab{b}}.

\bibitem[Wang et~al.(2022)Wang, Yang, Bremond, and Dantcheva]{wang2022latent}
Yaohui Wang, Di Yang, Francois Bremond, and Antitza Dantcheva.
\newblock Latent image animator: Learning to animate images via latent space navigation.
\newblock In \emph{ICLR}, 2022.

\bibitem[Wang et~al.(2023{\natexlab{a}})Wang, Chen, Ma, Zhou, Huang, Wang, Yang, He, Yu, Yang, et~al.]{lavie}
Yaohui Wang, Xinyuan Chen, Xin Ma, Shangchen Zhou, Ziqi Huang, Yi Wang, Ceyuan Yang, Yinan He, Jiashuo Yu, Peiqing Yang, et~al.
\newblock Lavie: High-quality video generation with cascaded latent diffusion models.
\newblock \emph{arXiv preprint arXiv:2309.15103}, 2023{\natexlab{a}}.

\bibitem[Wang et~al.(2023{\natexlab{b}})Wang, Ma, Chen, Dantcheva, Dai, and Qiao]{leo}
Yaohui Wang, Xin Ma, Xinyuan Chen, Antitza Dantcheva, Bo Dai, and Yu Qiao.
\newblock Leo: Generative latent image animator for human video synthesis.
\newblock \emph{arXiv preprint arXiv:2305.03989}, 2023{\natexlab{b}}.

\bibitem[Wang et~al.(2024)Wang, Yang, Bremond, and Dantcheva]{lia-pami}
Yaohui Wang, Di Yang, Francois Bremond, and Antitza Dantcheva.
\newblock Lia: Latent image animator.
\newblock \emph{IEEE Transactions on Pattern Analysis and Machine Intelligence}, pages 1--16, 2024.

\bibitem[Xie et~al.(2024)Xie, Xu, Song, Wang, Shi, and Luo]{x-portrait}
You Xie, Hongyi Xu, Guoxian Song, Chao Wang, Yichun Shi, and Linjie Luo.
\newblock X-portrait: Expressive portrait animation with hierarchical motion attention.
\newblock 2024.

\bibitem[Xu et~al.(2023)Xu, Song, Jiang, Zhang, Shi, Liu, Ma, Feng, and Luo]{xu2023omniavatar}
Hongyi Xu, Guoxian Song, Zihang Jiang, Jianfeng Zhang, Yichun Shi, Jing Liu, Wanchun Ma, Jiashi Feng, and Linjie Luo.
\newblock Omniavatar: Geometry-guided controllable 3d head synthesis.
\newblock In \emph{Proceedings of the IEEE/CVF Conference on Computer Vision and Pattern Recognition}, pages 12814--12824, 2023.

\bibitem[Yan et~al.(2021)Yan, Zhang, Abbeel, and Srinivas]{videogpt}
Wilson Yan, Yunzhi Zhang, Pieter Abbeel, and Aravind Srinivas.
\newblock Videogpt: Video generation using vq-vae and transformers.
\newblock \emph{arXiv preprint arXiv:2104.10157}, 2021.

\bibitem[Yang et~al.(2020)Yang, Zhu, Wu, Qian, Zhou, Zhou, and Loy]{transmomo2020}
Zhuoqian Yang, Wentao Zhu, Wayne Wu, Chen Qian, Qiang Zhou, Bolei Zhou, and Chen~Change Loy.
\newblock Transmomo: Invariance-driven unsupervised video motion retargeting.
\newblock In \emph{CVPR}, 2020.

\bibitem[Yu et~al.(2022)Yu, Tack, Mo, Kim, Kim, Ha, and Shin]{digan}
Sihyun Yu, Jihoon Tack, Sangwoo Mo, Hyunsu Kim, Junho Kim, Jung-Woo Ha, and Jinwoo Shin.
\newblock Generating videos with dynamics-aware implicit generative adversarial networks.
\newblock In \emph{ICLR}, 2022.

\bibitem[Yuan et~al.(2023)Yuan, Yong, Xiaodong, Fei, Yanbo, Xuan, Baoyuan, and Yujiu]{gong2023toontalker}
Gong Yuan, Zhang Yong, Cun Xiaodong, Yin Fei, Fan Yanbo, Wang Xuan, Wu Baoyuan, and Yang Yujiu.
\newblock Toontalker: Cross-domain face reenactment, 2023.

\bibitem[Zakharov et~al.(2019)Zakharov, Shysheya, Burkov, and Lempitsky]{zakharov2019few}
Egor Zakharov, Aliaksandra Shysheya, Egor Burkov, and Victor Lempitsky.
\newblock Few-shot adversarial learning of realistic neural talking head models.
\newblock In \emph{ICCV}, 2019.

\bibitem[Zhang et~al.(2024)Zhang, Wu, Liu, Zhao, Ran, Gu, Gao, and Shou]{show-1}
David~Junhao Zhang, Jay~Zhangjie Wu, Jia-Wei Liu, Rui Zhao, Lingmin Ran, Yuchao Gu, Difei Gao, and Mike~Zheng Shou.
\newblock Show-1: Marrying pixel and latent diffusion models for text-to-video generation.
\newblock \emph{International Journal of Computer Vision}, pages 1--15, 2024.

\bibitem[Zhang et~al.(2021)Zhang, Li, Ding, and Fan]{zhang2021flow}
Zhimeng Zhang, Lincheng Li, Yu Ding, and Changjie Fan.
\newblock Flow-guided one-shot talking face generation with a high-resolution audio-visual dataset.
\newblock In \emph{CVPR}, 2021.

\bibitem[Zhao and Zhang(2022)]{thinplate}
Jian Zhao and Hui Zhang.
\newblock Thin-plate spline motion model for image animation.
\newblock In \emph{CVPR}, 2022.

\end{thebibliography}
